\documentclass[conference]{IEEEtran}
\IEEEoverridecommandlockouts
\usepackage{cite}
\usepackage{amsmath,amssymb,amsfonts}
\usepackage{algorithmic}
\usepackage{graphicx}
\usepackage{textcomp}
\usepackage{xcolor}
\usepackage{multirow}
\usepackage{pdfpages}
\usepackage{graphicx}

\def\BibTeX{{\rm B\kern-.05em{\sc i\kern-.025em b}\kern-.08em
    T\kern-.1667em\lower.7ex\hbox{E}\kern-.125emX}}
\begin{document}

\title{EndoDepthL: Lightweight Endoscopic Monocular Depth Estimation with CNN-Transformer}

\author{
    \IEEEauthorblockN{Yangke Li}
    \IEEEauthorblockA{Imperial College London, London, United Kingdom}
    \IEEEauthorblockA{yangke.li23@imperial.ac.uk}
}

\maketitle

\begin{abstract}
In this study, we address the key challenges concerning the accuracy and effectiveness of depth estimation for endoscopic imaging, with a particular emphasis on real-time inference and the impact of light reflections. We propose a novel lightweight solution named EndoDepthL that integrates Convolutional Neural Networks (CNN) and Transformers to predict multi-scale depth maps. Our approach includes optimizing the network architecture, incorporating multi-scale dilated convolution, and a multi-channel attention mechanism. We also introduce a statistical confidence boundary mask to minimize the impact of reflective areas. To better evaluate the performance of monocular depth estimation in endoscopic imaging, we propose a novel complexity evaluation metric that considers network parameter size, floating-point operations, and inference frames per second. We comprehensively evaluate our proposed method and compare it with existing baseline solutions. The results demonstrate that EndoDepthL ensures depth estimation accuracy with a  lightweight structure.
\end{abstract}


\section{Introduction}

%monodepth在手术中的重要行，准确度，效率
%不能用LiDAR的原因，such as
%已有的monodepth自动驾驶领域方法不能直接搬过去，此外还包含独特挑战，最大挑战是反光

In contemporary surgical practices, monocular depth estimation is critical for endoscopic procedures, demanding both accuracy and efficiency \cite{stegemann2013fundamental}. Monocular cameras such as endoscopes are widely utilized but encounter the challenge of losing depth information from a 3D scene to a 2D image. This loss compels surgeons to rely heavily on their experience to discern the depth within the field of view, intensifying the complexity and decision-making pressure of the procedure \cite{stegemann2013fundamental}. While sensors like lidar could provide precise spatial positions, their integration into endoscopes is fraught with difficulties \cite{bernard2018deep}. The transition of monocular depth estimation methods such as Monodepth2\cite{godard2019digging} and LiteMono\cite{zhang2023lite} from the autonomous driving field is not straightforward, as endoscopic scenarios include unique challenges like inconsistent lighting. With the advancement in machine learning techniques, supervised learning has been explored to learn the 3D structure from 2D images \cite{saxena2005learning}. However, obtaining labelled endoscopic training data is prohibitively costly and challenging. Therefore, self-supervised learning, which extracts signals from image data without the need for additional labels, has gained increasing attention \cite{godard2017unsupervised,godard2019digging,gordon2019depth,wang2018learning}. Besides, reflections due to smooth organ surfaces and the real-time acquisition and computation of image data remain to be addressed, and maintaining a consistent frame rate is also critical. The unique combination of these challenges highlights the need for specialized depth estimation solutions in endoscopic applications.
\begin{figure}[]
\centering
\includegraphics[width=8.5cm]{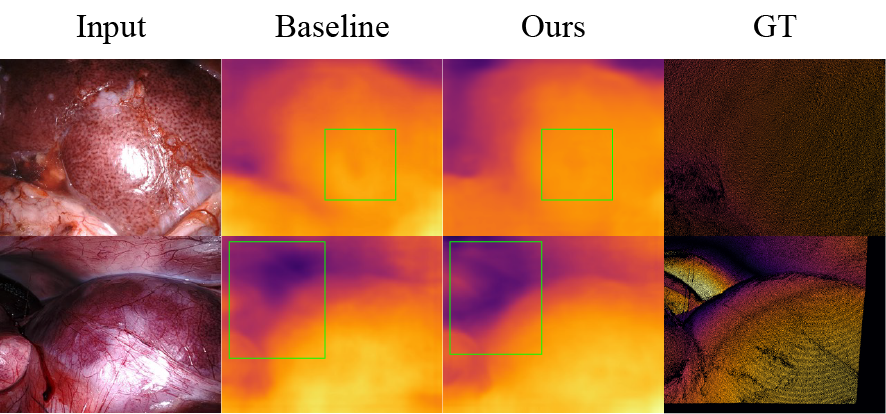}
\caption{ \textbf{Comparison of EndoDepthL with the baseline.} Our method effectively deals with the challenges in endoscopy, such as uneven lighting.}
\label{fig:picture001}
\end{figure}

%旋转矩阵改成外参数

The precision of depth estimation is crucial for endoscopic image analysis and surgery operations. Existing deep learning methods exhibit limitations in providing accurate and fast depth estimation results. Earlier solutions primarily hinged on Convolutional Neural Network(CNN), but the restrictions of the convolutional kernel make it hard to extract global features from images. Recently, numerous Transformer-based approaches have been proposed \cite{bae2023deep,guizilini2022multi,karpov2022exploring,liu2023self,mao2023bevscope,shi2023ega,zhang2023lite,varma2022transformers,zhao2022monovit,zhuang2023spdet}. While these methods effectively enhance the acquisition of global features, they also largely increase the network's parameter size, bringing down the inference speed significantly. This will affect the medical diagnosis results. Since algorithms must operate in real-time on edge devices like endoscopes, we cannot directly apply algorithms developed for specialized graphical computing devices to such equipment \cite{chen2019deep,wang2020deep}. Hence, there is an urgent to find a method to process endoscopic images in real-time while improving the accuracy and robustness of depth estimation, especially under strong light reflections.

%The principal objective of this work is to explore and propose a unique approach for addressing the real-time inference problem encountered during self-supervised monocular depth estimation in endoscopy. We aim to enhance the reliability and accuracy of endoscopic image processing by significantly increasing the inference speed while maintaining high depth estimation accuracy.

To this end, we propose a novel depth estimation method that concentrates on two main areas: firstly, we optimize the network architecture design to reduce the parameters of the network by designing a more lightweight network architecture; specifically, we incorporate multi-scale dilation convolution, and a multi-channel attention mechanism in the encoder to extract image features more efficiently. Moreover, we establish a confidence mask to minimize the influence of light-reflective regions on the training process and guide the network to focus on non-reflective regions.

Compared with existing endoscopic depth estimation methods, the contributions of our work can be summarized as follows:
\paragraph{Lightweight CNN-Transformer Encoder}
We design a method combining a Convolutional Neural Network (CNN) and a Transformer for predicting multi-scale depth maps from input images. This method combines dilated convolution with a cross-covariance attention mechanism to broaden the sensory field and capture global features without augmenting the network parameters. EndoDepthL ensures comparable performance to existing methods while facilitating faster inference speed.
\paragraph{Reflective Mask}
We propose a masking mechanism to address the common issue of reflections in endoscope environments. This mechanism effectively reduces the impact of reflective regions on depth estimation. Specifically, when calculating the loss function, the contribution of reflective regions is minimized to nearly zero. This enables EndoDepthL to prioritize depth estimation in non-reflective regions.
%and 达到sota
\paragraph{Complexity Evaluation}
We introduce the metric of parameter size, floating point operations and inference frames per second in our evaluation metrics to provide a more comprehensive evaluation of depth estimation over endoscopic images. To the best knowledge of the author, this is the first study in the field of endoscopic depth estimation that evaluates from the complexity perspective. We compare EndoDepthL with existing methods from both accuracy and efficiency perspectives, providing a benchmark for the practical application of depth estimation over endoscopic images.

In summary, our study provides a lightweight solution for endoscopic depth estimation that can mitigate the effect of reflections and is expected to further enhance the efficiency and safety of laparoscopic surgery. 
%This paper is structured as follows: Chapter 2 summarizes recent related studies, Chapter 3 describes our method in detail, Chapter 4 comprehensively evaluates the performance of our method, and Chapter 5 concludes the study.

\section{Related Work}
\begin{figure*}
\centering
\includegraphics[width=18cm]{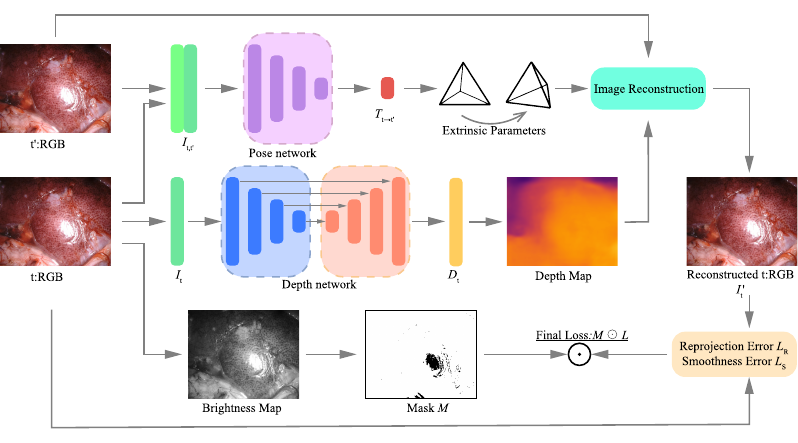}
\caption{ \textbf{Overview of the proposed method.} Put the source and target frames into the pose network and the target frame into the depth network. Each network extracts respective features: the pose network determines the transition from the source to the target, and the depth network produces initial depth predictions. Then reduces the reconstruction error by leveraging the camera's inherent parameters. To wrap up, a Statistical Confidence Boundary Mask is used to counteract the effects of light reflection, ensuring a more precise and stable result.}
\label{fig:picture001}
\end{figure*}

\subsection{Self-supervised Mono-Depth Estimation}
Self-supervised learning shows potential in depth estimation, with significant advancements due to monocular methods. Godard et al.\cite{godard2017unsupervised} introduced a self-supervised monocular network, using disparity for supervision. Zhou et al.\cite{zhou2017unsupervised} integrated multi-frame video sequences, estimating depth while learning camera motion.

However, challenges still exist in scenarios with dynamic scenes and changing lighting conditions. Zou et al.\cite{zou2018df} address these issues by introducing an unsupervised joint learning method of depth and flow called Df-net, which leverages cross-task consistency to model the relative motion within the scene. On the other hand, Li et al.\cite{li2018megadepth} tackle the same challenges by proposing Megadepth, a method that learns single-view depth prediction from a large and diverse set of Internet photos. There are extra issues with estimating the depth of small objects, as mentioned by Sattler et al.\cite{sattler2019understanding} and Wang et al.\cite{wang2020deep}.

Recent research addresses these issues from the loss function and neural network structures. Wang et al. \cite{wang2018occlusion} proposed an occlusion-aware loss function, while Guizilini et al.\cite{guizilini2020semantically}leveraged fixed pre-trained semantic segmentation networks to guide self-supervised representation learning via pixel-adaptive convolutions. There have been some efforts to enhance network structures for more efficient feature extraction. Zhao et al.\cite{zhao2019geometry} focused on geometric consistency to aid depth perception, integrating geometry-based constraints within their network structure. Yin et al.\cite{yin2019enforcing} solved the challenge by enforcing strong supervisory signals from the underlying 3D geometry, creating an alignment between monocular depth estimation and surface normals. Fu et al.\cite{fu2018deep} proposed a deep ordinal regression network, employing ordinal depth ranking among pixels to enable a more robust and discriminative representation of depth information. Additionally, Guizilini et al.\cite{guizilini20203d} introduced PackNet, a novel network structure that employs spatial packing and unpacking within convolutional layers. Similarly, Yang et al.\cite{yang2018lego} developed the LEGO (Learning Edge with Geometry all at Once) framework, which incorporates geometric constraints such as edges, planes, and vanishing points to improve depth estimation accuracy.

Transformer models could be beneficial in self-supervised monocular depth estimation. Vaswani et al.\cite{vaswani2017attention}has built based on the attention mechanism by Dosovitskiy et al.\cite{dosovitskiy2020image} for image recognition. Carion et al.\cite{carion2020end} proposed DETR, a Transformer-based object detection model. Transformers have shown potential in semantic segmentation\cite{zheng2021rethinking} and deep estimation\cite{cheng2021swin}. A recent work\cite{zhao2022monovit} combined plain convolutions with Transformer blocks to enhance local feature extraction and global information understanding in visual tasks.

Despite the potential advantages, Transformers still face challenges with high-resolution images\cite{katharopoulos2020transformers} and insufficient training data\cite{chen2020big}. These issues are particularly relevant for specific applications like endoscopic image depth estimation.

\subsection{Endoscopic Image Analysis}

Endoscopic image depth estimation is challenging due to distinct lighting conditions, complex backgrounds, and precision requirements. These difficulties have led to the emergence of various research strategies.

The unique lighting conditions in endoscopy affect image brightness and color, impacting depth estimation. Kohler et al. \cite{kohler2020laparoscopic} proposed a new color constancy method by separating spectral information of endoscopic images. Ma et al.\cite{ma2021structure} used Generative Adversarial Network (GAN) to enhance endoscopic images under inconsistent illumination.

Endoscopic images often contain complex backgrounds like blood and tissue debris, posing challenges to depth estimation since image noise affects the performance of depth estimation. To mitigate this, researchers have employed semantic segmentation techniques. Seo et al.\cite{seo2022semantic}and Zhu et al.\cite{zhu2019anatomynet} used deep learning to effectively separate regions of interest from complex medical images, hence improving the estimation accuracy in the following.

High-resolution images are necessary for precision in endoscopic surgery, demanding real-time, efficient depth estimation methods. Tang et al.\cite{howard2017mobilenets} proposed MobileNets, using depthwise separable convolution to improve model efficiency. Zhang et al.\cite{zhang2018shufflenet}introduced ShuffleNet, aiming to enhance model efficiency significantly.

Despite existing solutions providing several feasible ways to get depth estimation results, endoscopic image depth estimation remains an open question, requiring further work to improve accuracy and reliability.
\section{Methodology}
\begin{figure*}
\centering
%\includesvg[width=18.05cm]{fig222.svg}
\includegraphics[width=18.05cm]{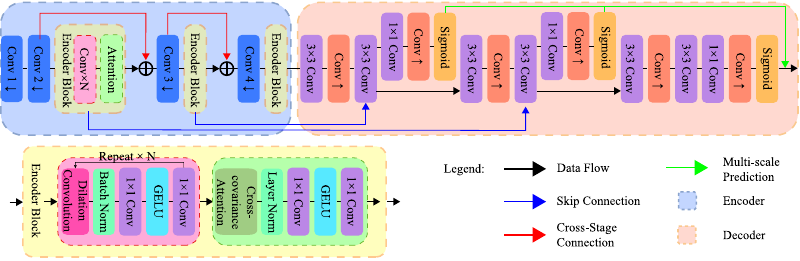}
\caption{ \textbf{Depth network.} We've enhanced feature extraction in the Encoder by incorporating an Encoder Block, consisting of convolution and attention components. We propose two Encoder network sizes(efficiency and performance) to meet varied requirements, as detailed in Table I.}
\label{fig:picture001}
\end{figure*}

\subsection{Self Supervised Loss Function}

EndoDepthL is based on a self-supervised principle, as illustrated in Fig. 2. Define two sequential images $I_{t}, I_{t+1}$, where $I_{t}$ is the reference frame and $I_{t+1}$ is the target frame, as described by \cite{godard2017unsupervised} and \cite{zhou2017unsupervised}. The depth network $D$ predicts the per-pixel depth value $d_{t}$ of the reference frame and converts it into a point cloud $X_{t}$:
\begin{equation}
X_{t} = d_{t}K^{-1}p_{t},
\end{equation}
where $K$ is the known camera intrinsic matrix, and $p_{t}$ is the pixel coordinates. The pose network $T$ predicts the relative pose from the reference frame to the target frame. Using this pose, we can transform the point cloud $X_{t}$ into the coordinate system of the target frame:
\begin{equation}
X_{t+1} = RX_{t} + t.
\end{equation}
The transformed point cloud $X_{t+1}$ is then projected back onto the image plane to obtain the pixel coordinates $p_{t+1}$ of the target frame:
\begin{equation}
p_{t+1} = KX_{t+1}.
\end{equation}
The reprojection pixel value $I_{t}^{'}$ is calculated using bilinear interpolation at the $p_{t+1}$ location of the target frame image $I_{t+1}$. The difference between this value and the original pixel value $I_{t}$ of the reference frame is used to calculate the reprojection error:
\begin{equation}
L_{R} = I_{t} - I_{t}^{'}.
\end{equation}
Minimizing this error allows the network to learn more accurate depth and relative pose during training.

Reconstruction uses both frames $I_{t-1}$ and $I_{t+1}$, and the smallest reconstruction error is selected for minimization:
\begin{equation}
L_{R} = min(L_{R,t-1}, L_{R,t+1}),
\end{equation}
where $L_{R,t-1}$ and $L_{R,t+1}$ are the reconstruction errors with $I_{t-1}$ and $I_{t+1}$, respectively.

Following \cite{godard2019digging}, a binary mask $\mu$ is defined to handle pixels that cannot be correctly projected:
\begin{equation}
\mu = min(L_{R,t-1}, L_{R,t+1}) < min(I_{t} - I_{t-1}, I_{t} - I_{t+1}),
\end{equation}
The associated loss function is:
\begin{equation}
L_{p} = \mu * L_{R},
\end{equation}

To encourage smoother depth map prediction, a smoothness loss based on the first-order derivatives of the image and depth map is added:
\begin{align}
L_{s} = |\partial_{x}d_{t}|e^{-|\partial_{x}I_{t}|} + |\partial_{y}d_{t}|e^{-|\partial_{y}I_{t}|},
\end{align}

The total loss function can be represented as:
\begin{equation}
L = L_{p} + \lambda L_{s},
\end{equation}
where $\lambda$ is an adjustable weight parameter.
In summary, the described method transforms unsupervised depth estimation into a process that jointly estimates the camera's depth and relative pose. This is based on the optimization of reprojection error, incorporation of a binary mask, and the application of a smoothness loss.

\begin{table}
\renewcommand{\arraystretch}{1.1} % 这行命令会将行高设为原来的1.5倍
\centering
\caption{Convolutional Parameters}
\begin{tabular}{c|cc}
\hline
              & \multicolumn{1}{c|}{Efficiency}        & Performance             \\ \hline
Input         & \multicolumn{2}{c}{320×256×3}                                    \\ \hline
cov1          & \multicolumn{1}{c|}{160×128×32}        & 160×128×64              \\ \hline
cov2          & \multicolumn{1}{c|}{80×64×32}          & 80×64×64                \\ \hline
dilation conv & \multicolumn{2}{c}{rate =1,2,3}                                  \\ \hline
cov3          & \multicolumn{1}{c|}{40×32×64}          & 40×32×128               \\ \hline
dilation conv & \multicolumn{2}{c}{rate =1,2,3}                                  \\ \hline
cov4          & \multicolumn{1}{c|}{20×16×128}         & 20×16×256               \\ \hline
dilation conv & \multicolumn{1}{c|}{rate =1,2,3,2,4,6} & rate =1,2,3,2,4,6,3,6,9 \\ \hline
\end{tabular}
\end{table}

\subsection{CNN-Transformer Lightweight Depth Network}

The proposed architecture comprises a DepthNet with Encoder-Decoder and a PoseNet with only Encoder. DepthNet, as illustrated in Fig.3, involves predicting multi-scale depth maps from an input image, while PoseNet is dedicated solely to predicting camera motion between sequential frames. Once these predictions are accomplished, a reconstructed target image is generated, and loss for model optimization is computed.

Traditional convolution operations are limited by their receptive field. To solve this, dilated convolution\cite{yu2015multi} is introduced into the model. This method expands the receptive field without extra parameters by interspersing gaps within the kernel elements, which can be formally represented as follows:
\begin{equation}
y[i] = \sum_{k=1}^{K} x[i + r \cdot k]w[k],
\end{equation}
where $w[k]$ refers to a filter of length $K$, and $r$ is the dilation rate. Dilated convolution's applied\cite{zhang2023lite} allows the model to grasp a broader contextual understanding and enhance the feature representation.

As shown in Table I, we set two sets of convolution parameters, corresponding to efficiency mode (Eff.) and performance mode (Perf.). Our model applies varying channel counts for downsampling convolutions, followed by three dilation convolutions with increasing rates. After the fourth downsampling, we use additional dilation convolutions with larger rates to obtain features in larger scales, totaling six iterations. In performance mode, we add three further dilation iterations.

 Inspired by the Transformer \cite{vaswani2017attention}, a global feature extraction method is employed that not only provides local features but also encompasses global information. Drawing from approaches like \cite{zhang2023lite}, this method utilizes a cross-covariance attention mechanism \cite{ali2021xcit}. It processes the attention between feature channels by linearly projecting the input feature map to derive the Query ($Q$), Key ($K$), and Value ($V$) components.The process can be depicted as:
\begin{equation}
\hat{X} = \text{Attention}(Q, K, V) + X,
\end{equation}
where
\begin{equation}
\text{Attention}(Q, K, V) = V \cdot \text{softmax}(Q^T \cdot K).
\end{equation}

\begin{table*}[]
\centering
\label{tab:result1}
\caption{Comparative Experimental Results}
\renewcommand{\arraystretch}{1.1} % 这行命令会将行高设为原来的1.5倍
\begin{tabular}{cc|ccccccc|ccc}
\hline
\multirow{2}{*}{Method} & \multirow{2}{*}{Data} & \multicolumn{7}{c|}{Accuracy}                                                                                                                      & \multicolumn{3}{c}{Complexity}                      \\ \cline{3-12} 
                        &                       & Abs Rel        & Sq Rel         & RMSE           & RMSE log       & $\delta$ \textless 1.25 & $\delta$ \textless $1.25^2$            & $\delta$ \textless $1.25^3$            & Param.     & FLOPs           & FPS \\ \hline
Monodepth2-Res18\cite{godard2019digging}        & M                     & 0.159          & 1.796          & 21.905         & 0.216          & 0.742            & 0.938                        & 0.999                        & 14.329M         & 8.038G          & 56.6            \\
Monodepth2-Res50\cite{godard2019digging}        & M                     & 0.143          & 1.466          & 21.589         & 0.191          & 0.757            & 0.981                        & \textbf{\textgreater{}0.999} & 32.522M         & 16.663G         & 32.82           \\
AF-SfMLearner\cite{shao2022self}           & M                     & 0.101          & 0.678          & 5.416          & 0.133          & 0.881            & \textbf{\textgreater{}0.999} & \textbf{\textgreater{}0.999} & 14.329M         & 5.359G          & 53.65           \\
LiteMono\cite{zhang2023lite}                & M                     & 0.133          & 1.225          & 7.332          & 0.167          & 0.791            & 0.993                        & \textbf{\textgreater{}0.999} & 3.069M          & 3.355G          & 60.9            \\
LiteMono-8m\cite{zhang2023lite}             & M                     & 0.112          & 0.96           & 6.556          & 0.135          & 0.888            & 0.998                        & \textbf{\textgreater{}0.999} & 8.766M          & 7.475G          & 45.67           \\
EndoDepthL-Eff.         & M                     & 0.104          & 0.727          & 5.38           & 0.135          & 0.883            & 0.998                        & \textbf{\textgreater{}0.999} & \textbf{2.143M} & \textbf{1.894G} & \textbf{62.8}   \\
EndoDepthL-Perf.        & M                     & \textbf{0.094} & \textbf{0.635} & \textbf{5.229} & \textbf{0.113} & \textbf{0.953}   & 0.998                        & \textbf{\textgreater{}0.999} & 10.882M         & 8.211G          & 45.01           \\ \hline
\end{tabular}
\end{table*}

\begin{table}[]
    \centering
    \caption{Metrics for Accurancy}
    \label{tab:metrics}
    \begin{tabular}{c|c}
        \hline
        Metric & Formula \\
        \hline
        Abs Rel & $\frac{1}{N} \sum\limits_{i} \frac{\left| d_{i} - \hat{d}_{i} \right|}{d_{i}}$ \\
        Sq Rel & $\frac{1}{N} \sum\limits_{i} \frac{\left( d_{i} - \hat{d}_{i} \right)^2}{d_{i}}$ \\
        RMSE & $\sqrt{\frac{1}{N} \sum\limits_{i} \left( d_{i} - \hat{d}_{i} \right)^2}$ \\
        RMSE log & $\sqrt{\frac{1}{N} \sum\limits_{i} \left( \log{d_{i}} - \log{\hat{d}_{i}} \right)^2}$ \\
        $\delta$ & $\frac{1}{N} \sum\limits_{i} \left[ \max \left( \frac{d_{i}}{\hat{d}_{i}}, \frac{\hat{d}_{i}}{d_{i}} \right) < \theta \right]$ \\
        \hline
    \end{tabular}
\end{table}

To further enhance the non-linearity of features, a GELU\cite{hendrycks2016gaussian} activation function is applied to the feature map. Then merged with the original input feature map to produce the final output feature map. Like the strategy proposed by\cite{he2016deep}, the enhanced feature map is combined with the original input features, leading to a richer feature map.

The following sections discuss a strategy to reduce the impact of reflections in depth estimation. We use a mask to help the model concentrate on key image areas. This approach ignores regions that reflections might distort.

\subsection{Statistical Confidence Boundary Mask}

% \usepackage{multirow}

Reflection and shadows may cause inconsistency in pixel intensities, violating the photometric consistency assumption. The masking mechanism can effectively mitigate this by excluding reflection areas from loss computation. 
The intensity map $L$ for an input image $I$ is computed:

\begin{equation}
L = 0.299 \cdot I_{r} + 0.587 \cdot I_{g} + 0.114 \cdot I_{b},
\end{equation}

where $I_{r}$, $I_{g}$, $I_{b}$ represent the red, green, blue channels of image $I$. A threshold $\tau$ distinguishes reflection from non-reflection areas. The intensity map $L$ is normalized to the [0, 1] interval, producing $L_{n}$:

\begin{equation}
L_{n} = \frac{L - \min(L)}{\max(L) - \min(L)},
\end{equation}

A mask $M$ is generated via a logistic function:

\begin{equation}
M = \frac{1}{1 + e^{-k \cdot (\tau-L_{n})}},
\end{equation}

where $k$ controls the transition smoothness. This mask transitions smoothly from 1 to 0 in reflection areas, whereas in non-reflection areas, values approach 1.

The loss function $L'$ incorporates the mask:

\begin{equation}
L' = M \odot L,
\end{equation}

where $\odot$ represents the Hadamard product. As reflection areas in the mask have values close to 0, their contribution to the loss function is negligible, effectively allowing the model to focus on non-reflective areas. The subsequent section presents experiments on a public dataset to validate these methods.

\section{Evaluation and Validation}

% \usepackage{multirow}

\subsection{Experiment Setup}

% \begin{table}[]
%     \centering
%     \caption{Training hyperparameters.}
%     \label{tab:hyperparameters}
%     \begin{tabular}{c|c}
%         \hline
%         \textbf{Parameter} & \textbf{Value} \\ \hline
%         Batch size & 8 \\ 
%         Weight decay & \(1 \times 10^{-2}\) \\ 
%         Learning rate & \(5 \times 10^{-4}\) \\
%         Epochs & 30 \\ \hline
%     \end{tabular}

% \end{table}

\paragraph{Dataset}
SCARED\cite{allan2021stereo} (Surgeries with CAmeras and Rigid Endoscopes Dataset) consist of endoscopic surgical videos collected using a da Vinci Xi endoscope and projector on fresh porcine cadaver abdominal anatomy to obtain high-quality depth maps. This process is performed at 5-10 different camera positions, following specific coded structured light imaging methods\cite{scharstein2014high}. The values in the depth maps are in millimeters, and invalid pixels are masked out. Since the camera must remain stationary during each structured light projection, the dataset is expanded with camera motion and warped depth maps using known camera poses from the da Vinci Xi kinematics. These poses are released as a 4x4 matrix, along with the stereo camera calibration for the sequence. The dataset is partitioned into training (15,351 frames), validation (1,705 frames), and testing sets (1,243 frames). The known intrinsic parameters of the endoscope guide the self-supervised training process, and during the evaluation phase, depth prediction remains within a 150mm constraint, simulating the physical limitations of endoscopic devices. Following \cite{godard2019digging}, we introduce data augmentation procedures, including horizontal flip, brightness, saturation, contrast adjustment, and hue jitter—with each occurring with a 50\% probability.

\begin{figure*}
\centering
\includegraphics[width=18.1cm]{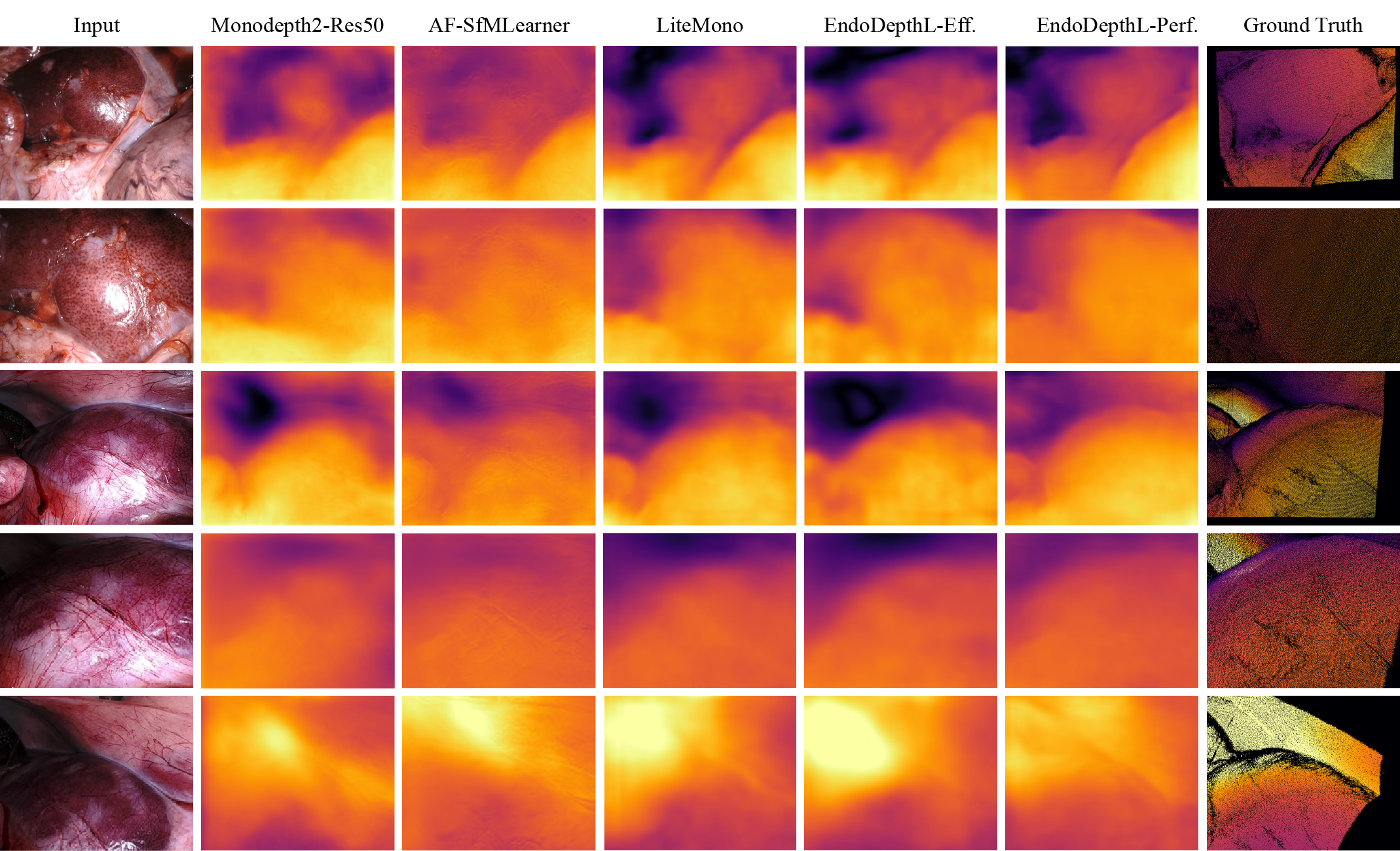}
\caption{ \textbf{Experimental result for our analysis.} We extracted representative frames from two distinct video segments. These carefully chosen frames encompass various perspectives, including frontal and lateral viewpoints, and capture different degrees of organ exposure. In some instances, the organs are fully visible, while in others, they are partially obscured or covered. From this figure, we can see that the EndoDepthL performance model is better with smoother and more accurate depth estimation.}
\label{fig:picture3}
\end{figure*}

%具体描述预处理过程，放在dataset部分

\paragraph{Hyperparameters}
The experiments were conducted on a system equipped with an 8-core CPU, 30GB of memory, and an NVIDIA T4 GPU, a mid-range unit resonating with the computational capabilities of edge devices. The system was hosted on Google Cloud and employed PyTorch version 1.12 for data processing and training. The training parameters include a batch size of 8 and the AdamW optimizer \cite{loshchilov2017decoupled}, with a weight decay of \(1 \times 10^{-2}\). The initial learning rate was set at \(5 \times 10^{-4}\), adhering to a cosine learning rate schedule. A monocular training session spanning 30 epochs took approximately 70 hours. The chosen configuration and methodological approach facilitated accurate modeling within the constraints of a medium-performance computational environment.

\paragraph{Evaluation Metrics}
The assessment employs standard monocular depth estimation metrics: Abs Rel, Sq Rel, RMSE, RMSE log, $\delta<1.25$, $\delta<1.25^{2}$and$\delta<1.25^{3}$. As shown in Table \ref{tab:metrics}, \( d_{i} \) represents the true depth values, \( \hat{d}_{i} \) denotes the predicted depth values, \( N \) is the total number of pixels, and \( \theta \) is a threshold. Additionally, we add an efficiency evaluation, considering the algorithm's overhead, including parameter size, floating point operations, and inference frames per second.

\subsection{Baseline Methods}
To highlight EndoDepthL's performance, we compared it with some popular baselines: Monodepth2, LiteMono, and AF-SfMLearner.

\textbf{Monodepth2}\cite{godard2019digging}, a typical classic method that includes versions ResNet18 and ResNet50, for benchmarking our study. Known for handling moving objects and occlusions, we followed its original parameter settings in experiments.

\textbf{Lite-Mono}\cite{zhang2023lite}, designed for autonomous driving challenges, blends CNN's local processing with Transformers' global capabilities. We adapted this method for endoscopic dataset, including fine-tuning the input size.

\textbf{AF-SfMLearner}\cite{shao2022self}, selected from open-source works, tackles endoscopic challenges such as inconsistent illumination. Its technique "Appearance Flow" aligns with our study's unique challenges.

\begin{table}[]
\centering
\caption{Ablation Study Result}
\renewcommand{\arraystretch}{1.1} % 这行命令会将行高设为原来的1.5倍
\begin{tabular}{c|ccccc}
\hline
Method           & AbsRel & SqRel & RMSE  & RMSElog & $\delta$\textless 1.25 \\ \hline
EndoDepthL-Eff.  & 0.104   & 0.727  & 5.380  & 0.135    & 0.883            \\
w/o mask         & 0.119   & 1.011  & 6.449 & 0.172    & 0.833            \\ \hline
EndoDepthL-Perf. & 0.094   & 0.635  & 5.229 & 0.113    & 0.953            \\
w/o mask         & 0.102   & 0.694  & 5.312 & 0.132    & 0.908            \\ \hline
\end{tabular}
\end{table}

\subsection{Experimental Results}

We compare EndoDepthL with baseline methods, and the results are listed in Table II and Fig. 4, covering performance and efficiency. ``M" denotes SCARED monocular videos. Our experiments are trained from scratch, and the best results are marked in bold.
%Inputs are reshaped to 320 by 256 这句话搬走
Our model demonstrates enhanced performance compared to the baseline methods. Specifically, it attains results akin to those of AF-SfMLearner but operates with reduced complexity. The efficiency of EndoDepthL allows for a 30\% reduction in network size while improving performance, or alternatively, it can decrease both the size and complexity by 3-5 times while maintaining similar levels of effectiveness.

We also compared with algorithms from autonomous driving, such as Litemono. EndoDepthL shows greater stability, which is essential for handling reflections on smooth organ surfaces in endoscopy. This stability leads to superior performance, with a four-fold increase in efficiency.

\subsection{Ablation Study}

We conducted an ablation study to validate the proposed mask module's effectiveness. The results are presented in Table IV and underscore the crucial function of this component. The experimental conditions were consistent with those of the previous comparative study.

The confidence mask is a vital part of our architecture, designed to diminish the impact of reflections during training. Without it, the reflections substantially affect the performance of EndoDepthL, increasing extreme values and having a noticeable influence on the Root Mean Square Error (RMSE). In our tests, removing the confidence mask led to an approximately 10\% increase in RMSE, accentuating the importance of managing reflections. Moreover, we observed a more pronounced decline in the processing ability of smaller, efficiency-oriented models when the mask was absent. This ablation study further substantiates the essential role of neutralizing reflection effects and establishes that our mask module is integral to improving the performance of lightweight networks.

%\textbf{Dilated Convolution Removal} When employing standard convolution instead of dilated convolution, performance slightly decreases. This suggests dilated convolution effectively enhances performance while maintaining the same complexity.

%\textbf{Covariance Attention Mechanism Removal} Although eliminating the attention mechanism decreases parameter count, it considerably downgrades performance. This underscores the importance of the attention mechanism in extracting global features.

\section{Conclusion}

This paper proposes a novel lightweight monocular depth estimation approach tailored for endoscopic applications. Utilizing a hybrid CNN and Transformer architecture, EndoDepthL adeptly extracts multi-scale local and global features from the endoscopic images. In addition, by integrating the confidence mask, EndoDepthL efficiently mitigates the detrimental effects of reflections, which is a standout challenge in endoscopic depth estimation. Experimental validation on the SCARED dataset demonstrates underscores our method's capability to balance low computational complexity with high estimation accuracy, paving the way for real-world deployment of depth estimation techniques in endoscopy.

\section*{Acknowledgment}
The author would like to thank Jiaping Xiao for the help and valuable feedback on this work.

\bibliographystyle{IEEEtran}
\bibliography{ref}

\end{document}